\documentclass[runningheads]{llncs}
\usepackage{amsmath}
\usepackage{mathtools}
\usepackage{graphicx}
\usepackage{graphics}
\usepackage{svg}
\usepackage{rotating}

\usepackage[colorinlistoftodos]{todonotes}
\usepackage[colorlinks=true, allcolors=blue]{hyperref}
\usepackage{caption}
\DeclareCaptionFont{mysize}{\fontsize{9}{9}\selectfont}
\usepackage[title]{appendix}
\usepackage{amsfonts}
\usepackage{multirow}
\usepackage{booktabs}
\usepackage[ruled,vlined,algo2e]{algorithm2e}

\newcommand{\R}{\mathbb{R}}
\DeclareMathOperator{\E}{\mathbb{E}}

\usepackage{acro}
\DeclareAcronym{PET}{
short=PET,
long=Positron Emission Tomography
}
\DeclareAcronym{SE}{
short=SE,
long=squeeze and excitation
}

\def\@fnsymbol#1{\ensuremath{\ifcase#1\or *\or \dagger\or \ddagger\or
   \mathsection\or \mathparagraph\or \|\or **\or \dagger\dagger
   \or \ddagger\ddagger \else\@ctrerr\fi}}
\newcommand{\ssymbol}[1]{^{\@fnsymbol{#1}}}

\begin{document}

\title{Anatomy-Constrained Contrastive Learning for Synthetic Segmentation without Ground-truth}

\author{Bo Zhou\inst{1}  \and Chi Liu\inst{1,2} \and James S. Duncan\inst{1,2}}
\institute{Biomedical Engineering, Yale University, New Haven, CT, USA \and Radiology and Biomedical Imaging, Yale University, New Haven, CT, USA}

\authorrunning{B. Zhou, etc} 
\titlerunning{Anatomy-Constrained Contrastive Learning} 

\maketitle              

\begin{abstract}
A large amount of manual segmentation is typically required to train a robust segmentation network so that it can segment objects of interest in a new imaging modality. The manual efforts can be alleviated if the manual segmentation in one imaging modality (e.g., CT) can be utilized to train a segmentation network in another imaging modality (e.g., CBCT/MRI/PET). In this work, we developed an anatomy-constrained contrastive synthetic segmentation network (AccSeg-Net) to train a segmentation network for a target imaging modality without using its ground-truth. Specifically, we proposed to use anatomy-constraint and patch contrastive learning to ensure the anatomy fidelity during the unsupervised adaptation, such that the segmentation network can be trained on the adapted image with correct anatomical structure/content. The training data for our AccSeg-Net consists of 1) imaging data paired with segmentation ground-truth in source modality, and 2) unpaired source and target modality imaging data. We demonstrated successful applications on CBCT, MRI, and PET imaging data, and showed superior segmentation performances as compared to previous methods. Our code is available at \href{https://github.com/bbbbbbzhou/AccSeg-Net}{https://github.com/bbbbbbzhou/AccSeg-Net}

\keywords{Contrastive Learning, Anatomy-Constraint, Synthetic Segmentation, Unsupervised Learning}
\end{abstract}
\acresetall

\section{Introduction}
Deep learning based image segmentation has wide applications in various medical imaging modalities. Over the recent years, numerous segmentation networks have been proposed to continuously improve the segmentation performance \cite{ronneberger2015u,oktay2018attention,roy2018concurrent,alom2019recurrent,isensee2020nnu,yu2020c2fnas}. These segmentation networks require training from large amounts of segmentation ground-truth on their target domain to achieve robust performance. However, a large amount of ground-truth data is not always available to several imaging modalities, such as intra-procedural CBCT, gadolinium-enhanced T1 MRI, and PET with different tracers, thus it is infeasible to directly obtain robust segmentation networks for them. In this work, we aim to obtain a robust segmenter on target modality without using target modality's ground-truth by leveraging the large amounts of source domain data (e.g., CT) with segmentation ground-truth.

Previous works on synthetic segmentation can be classified into two categories, including two-stage method \cite{zhang2018task} and end-to-end methods \cite{kamnitsas2017unsupervised,huo2018synseg}. Zhang et al. \cite{zhang2018task} developed a two-step strategy, called TD-GAN, for chest x-ray segmentation, where they first use a CycleGAN \cite{zhu2017unpaired} to adapt the target domain image to the domain with a well-trained segmenter, and then predict the segmentation on the adapted image. However, the segmentation performance relies on the image adaptation performance, thus the two-step process may prone to error aggregation. On the other hand, Kamnitsas et al. \cite{kamnitsas2017unsupervised} developed an end-to-end unsupervised domain adaptation for MRI brain lesion segmentation. However, they only used overlapping MRI modalities (e.g., FLAIR, T2, PD, MPRAGE) in both source and target imaging modalities to ensure performance. Later, Huo et al. \cite{huo2018synseg} proposed to directly concatenate CycleGAN and segmenter as an end-to-end network, called SynSeg-Net, and performed studies on two independent imaging modalities (e.g., MRI and CT). While SynSeg-Net achieved reasonable performance, there are still several issues. First, the training image of the segmentation network relies on high-quality adapted images from the CycleGAN part of SynSeg-Net. Without preserving anatomy structures and contents during the adaptation, the image could be adapted to a target domain image with incorrect structure/content, and negatively impact the subsequent segmentation network's training. Second, SynSeg-Net is a heavy design that relies on training 5 different networks simultaneously which requires careful hyper-parameter tuning, long training time, and high GPU memory consumption. 

To tackle these issues, we developed an anatomoy-constrained contrastive learning for synthetic segmentation network (AccSeg-Net), where we proposed to use anatomy-constraint and patch contrastive learning to preserve the structure and content while only using only 3 sub-networks. Given large amounts of CT data with segmentation ground-truth available from public dataset, we used CT as our source domain and validated our method's segmentation performance on target domains of CBCT, MRI, and PET. Our experimental results demonstrated that our AccSeg-Net achieved superior performance over previous methods. We also found our AccSeg-Net achieved better performance as compared to fully supervised methods trained with relatively limited amounts of ground-truth on the target domain. 

\begin{figure}[htb!]
\centering
\includegraphics[width=1.00\textwidth]{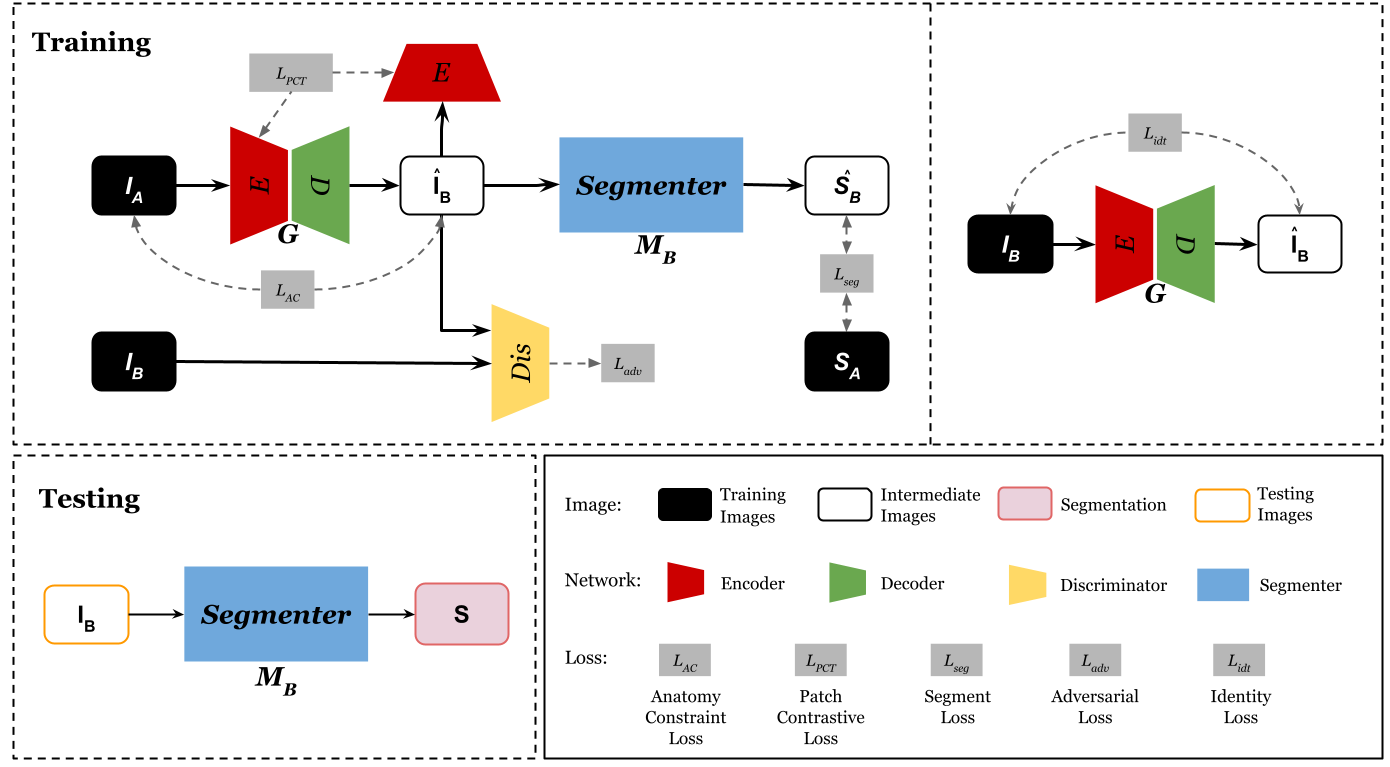}
\caption{The architecture of AccSeg-Net. Our AccSeg-Net consisting of a generator, a discriminator, and a segmented, is trained in an end-to-end fashion by 5 loss functions.}
\label{fig:network}
\end{figure}

\section{Methods}
Our AccSeg-Net includes two parts: 1) an anatomy-constraint contrastive translation network (Acc-Net), and 2) a segmentation network for the target domain segmentation. The architecture and training/test stages are shown in Figure \ref{fig:network}. Acc-Net aims to adapt images from domain $\mathbb{A}$ to domain $\mathbb{B}$. The anatomy-constraint loss and contrastive loss ensure structural information are not lost during the unpaired domain adaptation process, thus critical for training a robust segmenter in domain $\mathbb{B}$. On the other hand, the segmentation loss from the segmentation network is also back-propagated into the Acc-Net, providing addition supervision information for Acc-Net to synthesize image with correct organ delineation. More specifically, our AccSeg-Net contains a generator, a discriminator, and a segmenter. The generator $G$ adapts images from domain $\mathbb{A}$ to domain $\mathbb{B}$, the discriminator $D$ identifies real image from domain $\mathbb{B}$ or the adapted ones from $G$, and the segmenter $M_B$ predicts the segmentation $\hat{S_B}$ on adapted image from generator $G$. Training supervision comes from five sources: \\
\noindent\textbf{(a)} \textbf{adversarial loss} $\mathcal{L}_{adv}$ uses discriminator to minimize the perpetual difference between the generative image and the ground truth image in domain $\mathbb{B}$ by:
\begin{align}
    \mathcal{L}_{adv}(G,D,I_B,I_A)=\E_{I_B\sim\mathbb{B}}\left[ \log{D(I_B)} \right] + \E_{I_A\sim\mathbb{A}}\left[ \log{(1-D(G(I_A)))} \right]
\end{align}

\noindent\textbf{(b)} \textbf{identity loss} $\mathcal{L}_{idt}$ ensures the generator does not change target domain appearance when real sample in domain $\mathbb{B}$ is fed:
\begin{align}
    \mathcal{L}_{idt} = \E_{I_B\sim\mathbb{B}}\left[ \Vert G(I_B)-I_B \Vert_2^2 \right]
\end{align}

\noindent\textbf{(c)} \textbf{segmentation loss} $\mathcal{L}_{seg}$ is computed from the segmentation prediction from the adapted image $\hat{I_B}$ and the ground truth segmentation label from domain $\mathbb{A}$:
\begin{align}
    \mathcal{L}_{seg}=\E_{I_A\sim\mathbb{A}}\left[ 1 - \frac{2 | M_B(G(I_A)) \cap S_A |}{|M_B(G(I_A))| + |S_A|}  \right]
\end{align}

\begin{figure}[htb!]
\centering
\includegraphics[width=0.72\textwidth]{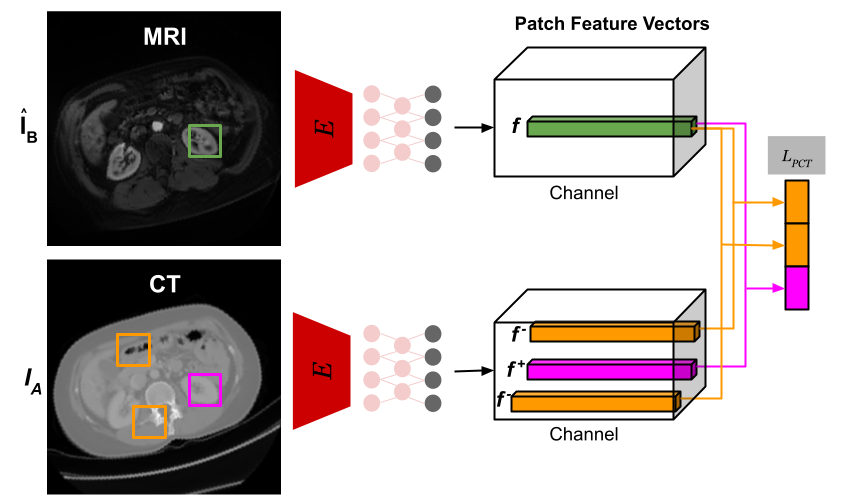}
\caption{Details of our patch contrastive loss ($\mathcal{L}_{PCT}$). The encoder (E) of the generator (G) followed by a fully-connected layer encodes the image patches into patch feature vectors. The $\mathcal{L}_{PCT}$ is computed using the patch feature vectors.}
\label{fig:pct}
\end{figure}

\noindent\textbf{(d)} \textbf{patch contrastive loss} $\mathcal{L}_{PCT}$ aims to associate the shared parts between the input image and the output image, while disassociating the non-shared parts. As illustrated in Figure \ref{fig:pct}, the kidney in the synthesized image $\hat{I}_B$ should have stronger association with the kidney in the input image $I_A$ than the other parts of the input image. We use the already trained encoder $E$ of generator $G$ (Figure \ref{fig:network}) to generate a feature stack of the image by choosing L layers of interest in E. Each spatial location within the individual level of the stack represents a patch of the input image \cite{chen2020simple}. In our implementation, each feature level goes through an additional fully-connected layer $H_l$. Thus, the feature stack can be formulated as $\{f_l\}_L = \{ H_l(E_l(I_A)) \}_L$, where $E_l$ is the l-th layer in E and $H_l$ is the full-connected layer. Denoting the number of spatial location in each layer as $s \in \{1,...,S_l\}$ and the number of feature channel in each layer as $C_l$, we can write the input associated feature as $f_l^s \in \R^{1 \times {C_l}}$ and the input disassociated feature as $f_l^{S \backslash s} \in \R^{{(S_l-1)} \times {C_l}}$. Similarly, the generator output's associated feature stack can be written as $\{\hat{f}_l\}_L = \{ H_l(E_l(G(I_A))) \}_L$. Then, we can compute our patch contrastive loss \cite{park2020contrastive} by:
\begin{align}
    \mathcal{L}_{PCT} = \E_{I_A\sim\mathbb{A}}\left[ \sum_{l=1}^L \sum_{s=1}^{S_l} \mathcal{L}_{CE}(\hat{f}_l^s, f_l^s, f_l^{S \backslash s} ) \right]
\end{align}
\begin{align}
    \mathcal{L}_{CE}(f,f^+,f^-) = -\log \left[ \frac{ \exp{(f \cdot f^+ / \alpha)} }{ \exp{(f \cdot f^+ / \alpha)} + \sum_{n=1}^N \exp{(f \cdot f^- / \alpha)}} \right]
\end{align}
where $\alpha$ is a temperature hyper-parameter for scaling the feature vector distance \cite{wu2018unsupervised}, and is empirically set to $\alpha=0.07$ here. The patches are randomly cropped during the training. 

\noindent\textbf{(e)} \textbf{anatomy-constraint loss} $\mathcal{L}_{AC}$ ensures the adaptation only alter the appearance of image while maintaining the anatomical structure, such that segmentation network $M_B$ can be trained correctly to recognize the anatomical content in the adapted image. We use both MIND loss \cite{yang2020unsupervised} and correlation coefficient (CC) loss to preserve the anatomy in our unpaired adaptation process:
\begin{equation}
    \mathcal{L}_{AC} = \lambda_{cc} \E_{I_A\sim\mathbb{A}}\left[
    \frac{Cov(G(I_A), I_A)}{\sigma_{G(I_A)} \; \sigma_{I_A}} \right] + \lambda_{mind} \E_{I_A\sim\mathbb{A}}\left[ \Vert F(G(I_A))-F(I_A) \Vert_1 \right]
\end{equation}
where the first term is the correlation coefficient loss. $Cov$ is the variance operator and $\sigma$ is the standard deviation operator. The second term is the MIND loss, and $F$ is a modal-independent feature extractor defined as $F_x(I) = \frac{1}{Z} exp \left( - \frac{K_x(I)}{V_x(I)} \right)$. Specifically, $K_x(I)$ is a distance vector of image patches around voxel $x$ with all the neighborhood patches within a non-local region in image $I$. $V_x(I)$ is the local variance at voxel x in image $I$. Here, dividing $K_x(I)$ with $V_x(I)$  reduce the influence of image modality and intensity range, and $Z$ is a normalization constant to ensure that the maximum element of $F_x$ equals to $1$. We set $\lambda_{cc}=\lambda_{mind}=1$ here to achieve a balanced training. 

Finally, the overall objective is a weighted combination of all loss listed above:
\begin{align}
    \mathcal{L}_{all} = \lambda_{1} \mathcal{L}_{adv} + \lambda_{2} \mathcal{L}_{idt} + \lambda_{3} \mathcal{L}_{seg} + \lambda_{4} \mathcal{L}_{PCT} + \lambda_{5} \mathcal{L}_{AC}    
\end{align}
where weighting parameters are set to $\lambda_{1} = \lambda_{2} = \lambda_{3} = \lambda_{4} = \lambda_{5} = 1$ to achieve a balanced training. We use a decoder-encoder network with 9 residual bottleneck for our generators, and a 3-layer CNN for our discriminators. For segmenter, we use a default setting of a 5-level UNet with concurrent SE module \cite{roy2018concurrent} concatenated to each level's output, named DuSEUNet. The segmenter in AccSeg-Net is interchangeable with other segmentation networks, such as R2UNet\cite{alom2019recurrent}, Attention-UNet\cite{oktay2018attention}, and UNet\cite{ronneberger2015u}.

\section{Experimental Results}
\noindent\textbf{Dataset Preparation} As liver segmentation is commonly available in CT domain, we chose CT as our source domain (domain $\mathbb{A}$), and aims to obtain CBCT/MRI/PET liver segmenters without their segmentation ground truth. For CT, we collected 131 and 20 CT volumes with liver segmentation from LiTS \cite{bilic2019liver} and CHAOS \cite{kavur2020chaos}, respectively. In the CBCT/MR domain, we collected 16 TACE patients with both intraprocedural CBCT and pre-operative MRI for our segmentation evaluations. In the PET domain, we collected 100 18F-FDG patients with abdominal scan. All the CBCT were acquired using a Philips C-arm system with a reconstructed image size of $384 \times 384 \times 297$ and voxel size of $0.65 \times 0.65 \times 0.65 mm^3$. The MRI and PET were acquired using different scanners with different spatial resolutions. Thus, we re-sampled all the CBCT, MR, PET and CT to an isotropic spatial resolution of $1 \times 1 \times 1 mm^3$. As a result, we obtained $13,241$ 2D CT images with liver segmentation, $3,792$ 2D CBCT images, $1,128$ 2D MR images, and $6,150$ 2D PET images. All the 2D images were resized to $256 \times 256$. With 16 CBCT and MRI patients in our dataset, we performed four-fold cross-validation with 12 patients used as training and 4 patients used as testing in each validation. For PET, we used 60 patients as training and 40 patients as testing, and qualitatively evaluated the results. Implementation details are summarized in our supplemental materials.

\begin{figure}[htb!]
\centering
\includegraphics[width=1.00\textwidth]{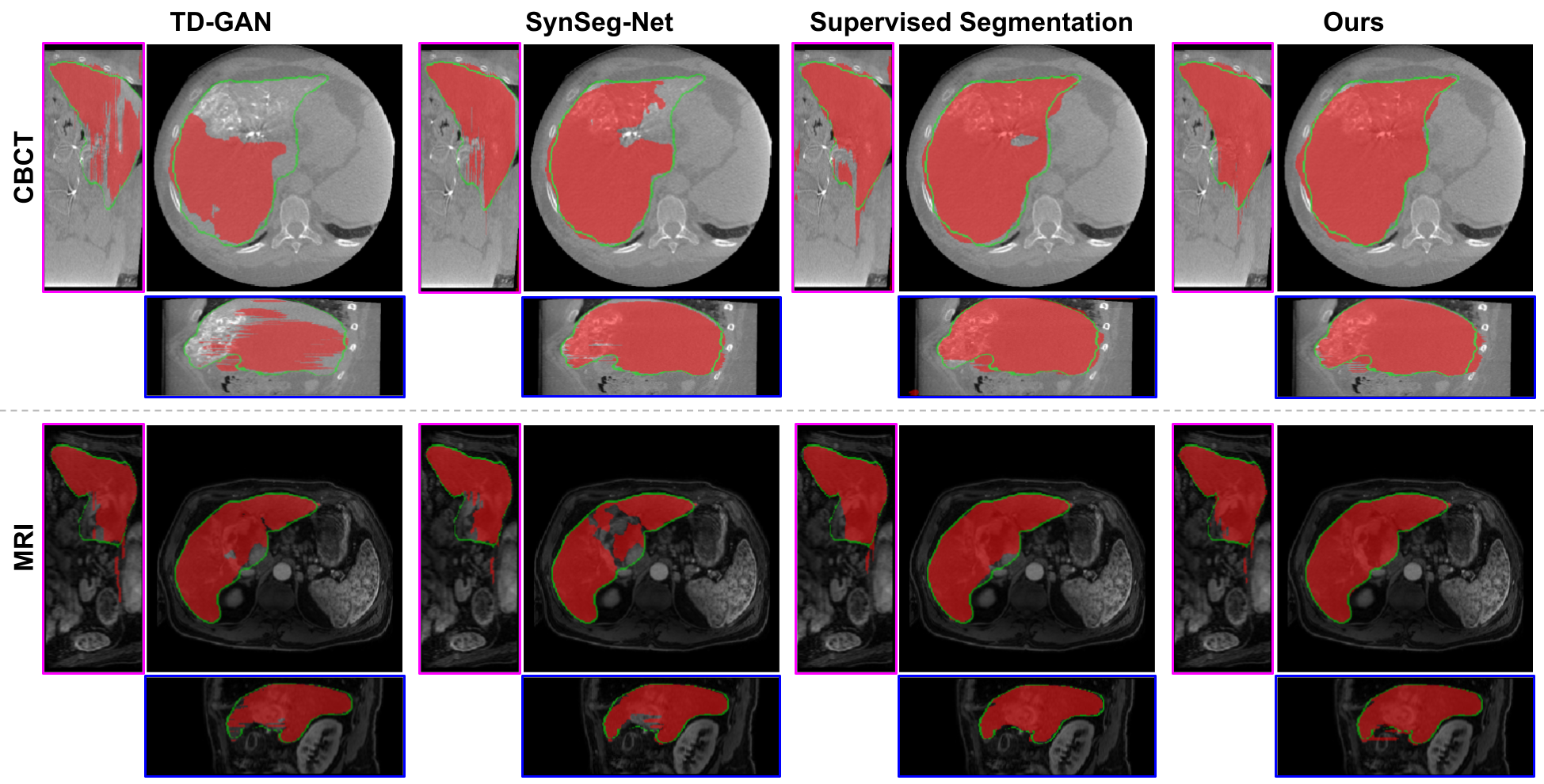}
\caption{Comparison of CBCT and MRI segmentation using different methods. Red mask: liver segmentation prediction. Green contour: liver segmentation ground truth. The coronal and saggital view are shown on left and bottom, respectively. }
\label{fig:compmethod}
\end{figure}

\begin{table} [htb!]
\footnotesize
\centering
\caption{Quantitative comparison of CBCT and MRI segmentation results using DSC and ASD(mm). * means supervised training with ground truth segmentation on the target domain (CBCT/MRI domain). -PCT means without patch contrastive learning. Best results are marked in \textcolor{blue}{blue}, and $\dagger$ means the difference between ours and SynSeg-Net are significant at $p<0.05$.}
\label{tab:compmethod}
    \begin{tabular}{|c|c|c|c||c|c|}
        \hline
        \textbf{CBCT}        & TD-GAN                  & SynSeg-Net              & Seg\textsubscript{CBCT}*          & Ours-PCT           & Ours             \\
        \hline
        Median DSC           & $0.685$                 & $0.870$                 & $0.882$                           & $0.827$            & \textcolor{blue}{$0.885$}          \\
        \hline
        Mean(Std) DSC        & $0.695(0.092)$          & $0.862(0.051)$          & $0.874(0.035)$                    & $0.831(0.079)$     & \textcolor{blue}{$0.893(0.029)^\dagger$}  \\
        \hline
        Median ASD           & $10.144$                & $7.289$                 & $9.190$                           & $9.824$            & \textcolor{blue}{$5.539$}          \\
        \hline
        Mean(Std) ASD        & $10.697(2.079)$         & $7.459(2.769)$          & $10.742(4.998)$                   & $10.381(3.739)$    & \textcolor{blue}{$5.620(1.352)^\dagger$}   \\
        \hline \hline
        \textbf{MRI}         & TD-GAN                  & SynSeg-Net              & Seg\textsubscript{MRI}*           & Ours-PCT           & Ours             \\
        \hline
        Median DSC           & $0.907$                 & $0.915$                 & $0.907$                           & $0.893$            & \textcolor{blue}{$0.920$}          \\
        \hline
        Mean(Std) DSC        & $0.900(0.044)$          & $0.912(0.029)$          & $0.859(0.102)$                    & $0.898(0.037)$     & \textcolor{blue}{$0.921(0.018)$}   \\
        \hline
        Median ASD           & $1.632$                 & $1.681$                 & $2.838$                           & $1.743$            & \textcolor{blue}{$1.468$}          \\
        \hline
        Mean(Std) ASD        & $2.328(2.070)$          & $1.916(1.142)$          & $3.660(2.522)$                    & $2.478(2.014)$     & \textcolor{blue}{$1.769(1.002)$}   \\
        \hline
    \end{tabular}
\end{table}

\noindent\textbf{Segmentation Results} After AccSeg-Nets are trained, we can extract the segmenters for liver segmentation on CBCT, MRI, and PET. We used Dice Similarity Coefficient (DSC) and Average Symmetric Surface Distance (ASD) to evaluate the quantitative segmentation performance. First, we compared our segmentation performance with TD-GAN\cite{zhang2018task}, SynSeg-Net\cite{huo2018synseg}, and segmenter directly supervised trained on target domain images with limited liver annotations (Seg\textsubscript{CBCT} / Seg\textsubscript{MRI}). For a fair comparison, we used DuSEUNet as segmentation network in all the compared methods. The visual comparison is shown in Figure \ref{fig:compmethod}, and the quantitative comparison is summarized in Table \ref{tab:compmethod}. As we can observe, the image quality of CBCT is degraded by metal artifacts and low CNR. TD-GAN's results are non-ideal as it requires adapting the input CBCT to CT first, and the segmentation relies on the translated image quality. The unpaired and unconstrained adaption from CBCT to CT is challenging as it consists of metal artifact removal and liver boundary enhancement. The multi-stage inference in TD-GAN thus prone to aggregate prediction errors into the final segmentation. SynSeg-Net with single-stage segmentation help mitigate the prediction error aggregation, but the segmentation is still non-ideal due to lack of structure and content constraint in the unpaired adaptation. On the other hand, our AccSeg-Net with patch contrastive learning and anatomy constraint achieved the best segmentation results. Compared to the segmenters trained on target domains using relatively limited annotation data (12 CBCT/MRI patients), our AccSeg-Net trained from large-scale conventional CT data (151 patients) can also provide slightly better segmenters.

\begin{figure}[htb!]
\centering
\includegraphics[width=1.00\textwidth]{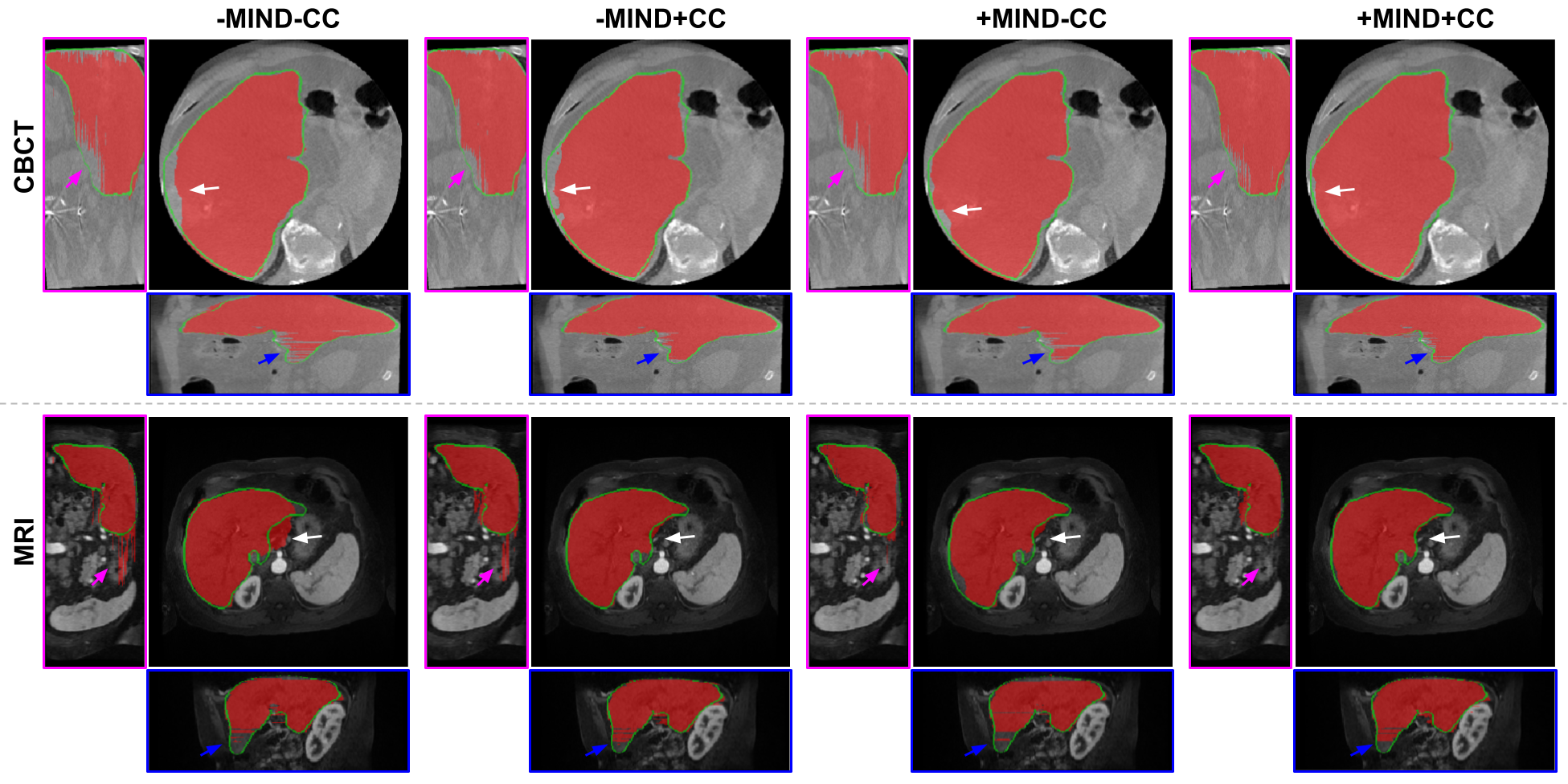}
\caption{Comparison of CBCT and MRI segmentation using different anatomy-constraint settings. Red mask: liver segmentation prediction. Green contour: liver segmentation ground truth.}
\label{fig:compap}
\end{figure}

\begin{table} [htb!]
\footnotesize
\centering
\caption{Quantitative comparison of CBCT and MRI segmentation results using different anatomy-constraint settings in AccSeg-Net. -MIND and -CC means without MIND loss and without CC loss, respectively. Best results are marked in \textcolor{blue}{blue}. }
\label{tab:compap}
    \begin{tabular}{|c|c|c|c|c|}
        \hline
        \textbf{CBCT}        & -MIND-CC             & +MIND-CC             & -MIND+CC             & +MIND+CC  \\
        \hline
        Mean(Std) DSC        & $0.869(0.059)$       & $0.883(0.041)$       & $0.885(0.039)$       & \textcolor{blue}{$0.893(0.029)$}                     \\
        \hline
        Mean(Std) ASD        & $7.397(2.268)$       & $6.648(2.147)$       & $6.424(2.110)$       & \textcolor{blue}{$5.620(1.352)$}                     \\
        \hline \hline
        \textbf{MRI}         & -MIND-CC             & +MIND-CC             & -MIND+CC             & +MIND+CC  \\
        \hline
        Mean(Std) DSC        & $0.913(0.030)$       & $0.916(0.024)$       & $0.918(0.020)$       & \textcolor{blue}{$0.921(0.018)$}                      \\
        \hline
        Mean(Std) ASD        & $1.923(1.153)$       & $1.886(1.138)$       & $1.803(1.109)$       & \textcolor{blue}{$1.769(1.002)$}                      \\
        \hline
    \end{tabular}
\end{table}

\begin{figure}[htb!]
\centering
\includegraphics[width=0.99\textwidth]{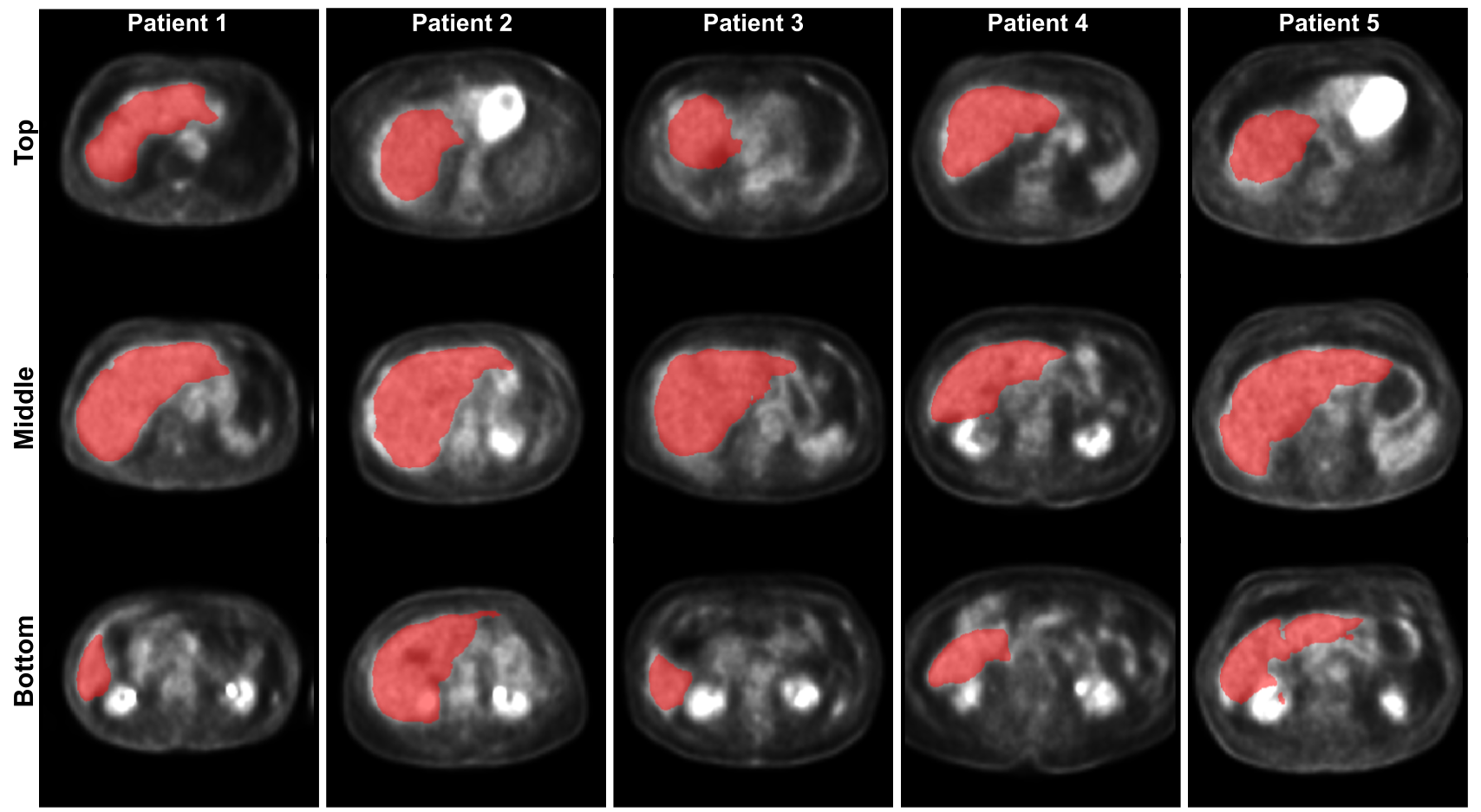}
\caption{PET liver segmentation prediction on 5 different patients at three different latitudes.}
\label{fig:pet}
\end{figure}

Then, we analyzed the effect of using different anatomy-constraint settings in our AccSeg-Net. The results are visualized in Figure \ref{fig:compap}, and summarized in Table \ref{tab:compap}. As we can see, adding either MIND loss or CC loss help improve our segmentation performance, while combining both anatomy constraint losses yields the best segmentation performance of our AccSeg-Net. We also performed ablation studies on using different segmentation networks in our AccSeg-Net, including R2UNet\cite{alom2019recurrent}, Attention-UNet\cite{oktay2018attention}, and UNet\cite{ronneberger2015u}. The visualization and quantitative evaluation are summaried in our supplemental materials. We observed that AccSeg-Net can be adapted to different segmentation networks, and yields reasonable segmentation results. Additional PET liver segmentation results from our AccSeg-Net are shown in Figure \ref{fig:pet} for visual evaluation. Our AccSeg-Net can also provide reasonable segmentation on PET data without using any ground-truth annotation in PET domain. 

\section{Discussion and Conclusion}
In this work, we proposed a novel framework, called AccSeg-Net, for synthetic segmentation without the target domain's ground truth. Specifically, we proposed to use anatomy-constraint and patch contrastive learning in our AccSeg-Net to ensure the anatomy fidelity during the unsupervised adaptation, such that the segmenter can be trained on the adapted image with correct anatomical contents. We demonstrated successful applications on CBCT, MRI, and PET imaging data, and showed superior segmentation performances as compared to previous methods. The presented work also has potential limitations. First, our segmentation performance is far from perfect, and extending our framework to 3D and with enhanced segmentation loss may further improve the performance. Moreover, our method is an open framework, and substituting the segmentation network with a more advanced network may also help improve the performance. In our future work, we will also explore diseases and multi-organ segmentation on other imaging modalities, such as ultrasound and radiography.

\bibliographystyle{splncs}
\bibliography{bibliography}

\end{document}